# A MULTI-STAGE FRAMEWORK WITH CONTEXT INFORMATION FUSION STRUCTURE FOR SKIN LESION SEGMENTATION

*Yujiao Tang[1], Feng Yang[1], Shaofeng Yuan [1, 2], Chang'an Zhan[1]*

[1]Guangdong Provincial Key Laboratory for Medical Image Processing, School of Biomedical Engineering, Southern Medical University, Guangzhou, China
[2]Shanghai United Imaging Healthcare Co. Ltd., Shanghai, China

**ABSTRACT**

The computer-aided diagnosis (CAD) systems can highly improve the reliability and efficiency of melanoma recognition. As a crucial step of CAD, skin lesion segmentation has the unsatisfactory accuracy in existing methods due to large variability in lesion appearance and artifacts. In this work, we propose a framework employing multi-stage UNets (MS-UNet) in the auto-context scheme to segment skin lesion accurately end-to-end. We apply two approaches to boost the performance of MS-UNet. First, UNet is coupled with a context information fusion structure (CIFS) to integrate the low-level and context information in the multi-scale feature space. Second, to alleviate the gradient vanishing problem, we use deep supervision mechanism through supervising MS-UNet by minimizing a weighted Jaccard distance loss function. Four out of five commonly used performance metrics, including Jaccard index and Dice coefficient, show that our approach outperforms the state-of-the-art deep learning based methods on the ISBI 2016 Skin Lesion Challenge dataset.

*Index Terms*— skin lesion segmentation, UNet, auto-context scheme, context information fusion structure, deep supervision mechanism

## 1. INTRODUCTION

Melanoma has been considered as the most deadly type of skin cancer, and the estimated number of new cancer cases has reached 91,270 in 2018 in the United States [1]. Early diagnosis is of great importance for increasing 5-year survival rate [2]. Identifying melanoma by dermatologists has the poor reproducibility [3]. Meanwhile, the diagnostic accuracy is reduced due to a number of reasons, such as irregular and fuzzy lesion borders, varying shapes, color and sizes, low contrast between lesion and surrounding skin, and a variety of artifacts. Therefore, the computer-aided diagnosis systems for melanoma recognition are of high demand.

This research was supported by the National Natural Science Foundation of China (No.: 61771233). Feng Yang is the corresponding author (e-mail: yangf@smu.edu.cn).

Accurate segmentation of skin lesion is a basic step in computerized analysis system. Many classical algorithms have been developed for this challenging task, such as thresholding, region based approaches, edge based methods and deformable models [4]. Recently, convolutional neural networks have achieved amazing performance in computer vision tasks. Among them, fully convolutional networks (FCN) proposed by Long *et al.* [5] were widely used in semantic segmentation. FCN predicts pixel-wise labels by replacing fully-connected layers with $1 \times 1$ convolutional layers. Yuan *et al.* [6] proposed a 19-layer FCN with Jaccard distance based loss function to segment skin lesion and evaluated the model on the ISBI 2016 Skin Lesion Challenge dataset as well. Bi *et al.* [7] presented a multi-stage FCN (mFCN) approach for accurate skin lesion segmentation. Then, they employed a parallel integration method based on cellular automata to combine the outputs of every stage. In 2015, Ronneberger *et al*. [8] proposed the UNet, an encoder-decoder FCN for automatic biomedical image segmentation with the limited training data. In UNet, the skip connection architecture combines high resolution feature from shallow layers with upsampled feature from deep layers in order to yield more precise segmentation. Due to its high efficiency and accuracy, UNet is widely applied to medical image analysis and medical imaging.

Inspired by Bi *et al.* [7], we propose a framework with multi-stage UNets (MS-UNet) and a weighted Jaccard distance loss for automatic detection of the lesion's border. The architecture of MS-UNet model with deeply-supervised (DS) learning strategy is shown in Fig. 1. Specially, we inject multiple UNets into the auto-context scheme [9] to segment skin lesion without any pre- or post-processing and train it end-to-end. In MS-UNet, a context information fusion structure (CIFS) is presented to integrate the low-level and context information in the multi-scale feature space. In order to alleviate the training difficulties caused by gradient vanishing problem, we supervise MS-UNet by minimizing a weighted loss function in spirit of deep supervision mechanism. We evaluate our MS-UNet on the ISBI 2016 Skin Lesion Challenge dataset and examine the utility of CIFS within the framework. In addition, we compare our approach with competitive methods.

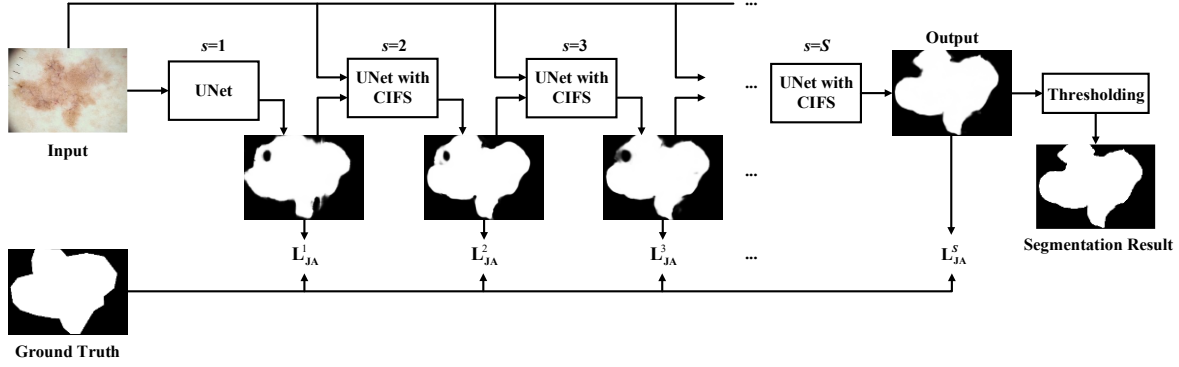

**Fig. 1**. The architecture of MS-UNet with deeply-supervised learning strategy.

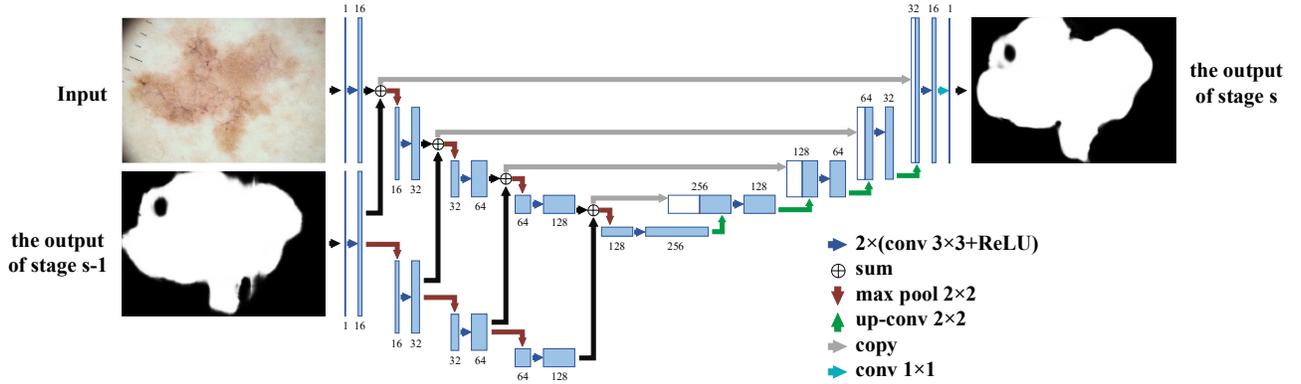

**Fig. 2**. UNet with context information fusion structure (CIFS). CIFS encodes and incorporates the context information from the output of stage *s*-1 into the *s*-th UNet in the multi-scale feature space.

## 2. METHODS

The overview of the proposed MS-UNet model is shown in Fig. 1. We first use the UNet which is appropriate for biomedical image segmentation to produce a coarse probability map. In each of the following UNet, we apply the context information fusion structure (CIFS), illustrated in Fig. 2, to incorporate the context information of the probability map generated by previous stage into the encoder path of UNet in the multi-scale feature space. The UNet with CIFS gains more accurate result than preceding one. Considering the training difficulties, we adopt the deep supervision strategy in the MS-UNet framework and the Jaccard distance loss helps the model to supervise every stage. Finally, we employ a simple thresholding technique to the output of the last stage and obtain the segmentation result.

### 2.1. Multi-stage UNet architecture

Facing the challenging task of skin lesion segmentation, we take advantage of multiple UNets in an auto-context scheme to improve the segmentation performance. Once n dermoscopic images $X = \{x_1, x_2, \ldots, x_n\}$ are fed into the MS-UNet, a set of probability maps $Y = \{Y^1, Y^2, \ldots, Y^S\}$ are obtained. We choose the thresholded $Y^S = \{y_1^S, y_2^S, \ldots, y_n^S\}$ as the final segmentation result, where $S$ is the number of UNet and modified UNet (see Fig. 2). The iterative process of multi-stage architecture is illustrated as:

$$\begin{cases} Y^s = D(E(X; W_e^s); W_d^s) & s = 1 \\ Y^s = D(A(X; W_{e1}^s, W_{e2}^s); W_d^s) & s > 1 \end{cases} \quad (1)$$

where $W_e^s$ and $W_d^s$ are the parameters which are learned in encoding process $E$ and decoding process $D$ of stage $s$ ($s = 1, 2, \ldots, S$), $W_{e1}^s$ and $W_{e2}^s$ represent the parameters which are optimized in the two parallel encoding processes of *s*-th stage of UNet with CIFS, $Y_1^s$ is the output of stage $s$ and $A$ is the encoder with CIFS.

In the regular auto-context scheme, the multiple basic models are typically pre-trained separately and then these models are fine-tuned on dermoscopic images, which causes mismatches between adjacent stages. Our multi-stage architecture MS-UNet is trained in an end-to-end manner and then the segmentation of skin lesion is directly performed. By doing this, more room is given for the model to adjust automatically according to the RGB images and the adjacent stages fit better.

### 2.2. Context information fusion structure

In the classic auto-context algorithm [9], the probability map generated by each classifier is concatenated with the RGB image and then they are fed into the next classifier. Thus this model is capable of integrating high-level context and low-level appearance information in the image space.

For some structured problems, information fusion in the feature space may make model understand the full image completely. In this work, we add context information fusion structure (CIFS) into UNet from the second stage on. The CIFS integrates the low-level and context information in the multi-scale feature space to correct the mistakes in the encoding process. As demonstrated in Fig. 2, RGB dermoscopic image and output of the previous stage are treated the same on both encoding paths, each of which has the repetitive step which contains two 3 × 3 convolutions followed by a rectified linear unit (ReLU) and a 2 × 2 max pooling to capture abstract semantic features. Before every max pooling operation, we adopt addition of corresponding elements to fuse features of two paths. By doing this, the proposed model achieves information integration in multi-level. Like the UNet, the skip connection architecture bypasses the fused features from the encoder to decoder to reduce information loss. In decoder path, a 2 × 2 deconvolution and two 3 × 3 convolutions followed by a ReLU form a building block and it is repeated to restore the image size. Then, a 1 × 1 convolution is appended to predict pixel-wise probability.

**2.3. Deep supervision mechanism**

In deep neural networks, vanishing gradient makes optimization difficult and brings a series of problems in training [10-11], such as decrease of convergence rate and reduction of discrimination capability of model. To alleviate these problems, deep supervision mechanism [11] is employed. As shown in Fig. 1, an auxiliary supervision is applied to each stage before the last UNet. We choose Jaccard distance [5] as the basic loss function. The MS-UNet obtains the segmentation result by classifying each pixel of input, so the Jaccard distance loss function of stage $s$ is computed using Eq. (2). Finally, the MS-UNet is deeply supervised by minimizing an integrated loss function $L$ defined in Eq. (3). In Eq. (2), $t_{ij}^s$ and $p_{ij}^s$ are the values of ground truth and predicted probability, respectively. In Eq. (3), $\alpha^s$ represents the weight of Jaccard distance loss function of stage $s$ ($s = 1, 2, …, S$).

$$L_{JA}^s = 1 - \frac{\sum_{i,j}(t_{ij}^s p_{ij}^s)}{\sum_{i,j}(t_{ij}^s)^2 + \sum_{i,j}(p_{ij}^s)^2 - \sum_{i,j}(t_{ij}^s p_{ij}^s)} \quad (2)$$

$$L = \sum_{s}^{S} \alpha^s L_{JA}^s \quad (3)$$

## 3. EXPERIMENTAL RESULTS

**3.1. Data description and implementation details**

The MU-Net is estimated on a public challenge dataset, which is supplied for the ISBI 2016 challenge named "Skin Lesion Analysis Towards Melanoma Detection" [12]. There are 900 training images and 379 testing images in this dataset. All images are annotated by experts and the size ranges from 1022 × 767 to 4288 × 4288. Before training, images were resized to 224 × 160 and then normalized in image-wise level. For the data augmentation, we randomly flipped images horizontally and vertically, then rotated them in an angle range (-25º, 25º). We chose five evaluation metrics [12] from the ISBI 2016 challenge to estimate the proposed method, including: Jaccard index (JA), Dice coefficient (DI), sensitivity (SE), specificity (SP) and accuracy (AC).

Considering segmentation accuracy and computation cost, we reduced the number of feature channels to a quarter of its original size at each step of encoder and decoder parts. The numbers of feature maps in a single UNet are shown in Fig. 2. Our model employs 4 UNets for fair comparison to mFCN-PI [7] and it will perform better if more stages involved. We set weights of Jaccard distance loss for 4 stages as 0.7, 0.8, 0.9 and 1 empirically. The weighted loss is minimized by the Adam optimizer with an initial learning rate 0.0001. For the output of the last stage, we applied thresholding with a value of 0.7 to obtain the precise segmentation result. We implemented the proposed method using a NVIDIA GTX1080Ti GPU with 11GB VRAM in Keras library and TensorFlow framework. The model takes 16 hours to train over 120 epochs with a mini batch size of 16.

**3.2. Results**

As shown in Table 1, both CIFS in Section 2.2 and DS learning strategy in Section 2.3 help improve the performance of MS-UNet model on all evaluation metrics. Note that the proposed CIFS in Fig. 2 is slightly better than the simple concatenation operation in the original auto-context algorithm (row 1 vs. row 2 and row 3 vs. row 4). Without and with DS, CIFS improves scores of JA metric by 0.56 and 0.37, respectively, in comparison to concatenation of RGB image and probability map. We conjecture that information fusion in the multi-scale feature space can capture more abstract and complex features for the dense prediction problems. Deep supervision mechanism further improves segmentation performance in both CIFS and concatenation conditions by withstanding the gradient vanishing. In Table 1, we also compared the MS-UNet model with top 3 methods on ISCI 2016 challenge leaderboard (ExB, CUMED [13] and Rahman), and other the-state-of-art approaches such as mFCN-PI [7], GBLB [14], Fus-CNN [15], SA-FCN [16] and DAC-FCN [17]. From the quantitative results, our model outperforms all methods in JA, DI, SE and AC metrics. Compared to the competitive mFCN-PI method, MS-UNet raises JA by 0.70 higher and AC from 95.51 to 95.87. The modified UNet with CIFS allows the high-level context information to be processed again in the multi-scale feature space, which makes MS-UNet more discriminative to skin lesion.

**Table 1.** Performance of MS-UNet, top 3 methods on ISIC 2016 challenge leaderboard and other state-of-the-art approaches. Bold numbers indicate the best performance.

| Method | JA | DI | SE | SP | AC |
|---|---|---|---|---|---|
| MSUN+C | 83.22 | 89.91 | 91.36 | 95.98 | 95.34 |
| MSUN+CIFS | 83.78 | 90.43 | 91.50 | 96.06 | 95.43 |
| MSUN+C+DS | 84.97 | 91.13 | 92.53 | 95.86 | 95.72 |
| MSUN+CIFS+DS | **85.34** | **91.47** | **92.67** | 96.42 | **95.87** |
| ExB | 84.30 | 91.00 | 91.00 | 96.50 | 95.30 |
| CUMED [12] | 82.90 | 89.70 | 91.10 | 95.70 | 94.90 |
| Rahman | 82.22 | 89.50 | 88.00 | 96.90 | 95.20 |
| mFCN-PI [6] | 84.64 | 91.18 | 92.17 | 96.54 | 95.51 |
| GBLB [13] | 84.10 | 90.70 | 93.80 | 95.20 | 95.30 |
| Fus-CNN [14] | 82.90 | 90.65 | / | / | / |
| SA-FCN [15] | 83.36 | 90.24 | 91.82 | 95.23 | 95.10 |
| DAC-FCN [16] | 83.30 | 90.11 | 90.15 | **97.00** | 95.12 |

MSUN denotes MS-UNet, C denotes concatenation, CIFS denotes context information fusion structure, DS denotes deep supervision.

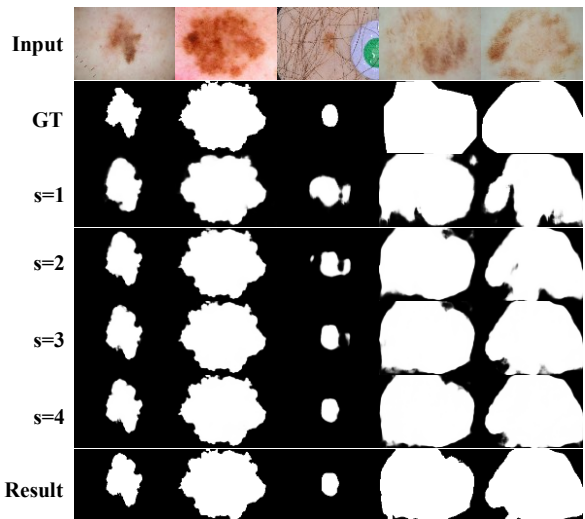

**Fig. 3**. Qualitative segmentation results of MS-UNet. Row 1: input images. Row 2: ground truths. Row 3~6: the outputs of 4 stages. Row 7: final segmentation results.

Fig. 3 presents four qualitative results. In the first two columns, the borders of lesion region are highly irregular. It's clear that the segmentation results have been refined stage to stage. As observed in the 3$^{rd}$ column, the interference of hair in the background is more and more suppressed with the increase of $s$. For images of low contrast as in the examples in the last two columns, the $s$-th probability map is much closer to the ground truth than previous stages.

## 4. CONCLUSION

We propose the multi-stage UNets (MS-UNet), which stacks multiple UNets in an auto-context scheme, to segment skin lesions. In this model, we feed the probability map produced by each UNet as well as the RGB image into the next stage of UNet. From the second stage on, a context information fusion structure (CIFS) is added to encoder to integrate the low-level and context information in the multi-scale feature space. Each stage is supervised by minimizing a Jaccard distance loss in combat with the gradient vanishing problem. The experiments demonstrate that the MS-UNet model achieves a state-of-the-art segmentation result on the ISCI 2016 dataset. In future work, we will evaluate the MS-UNet model and the proposed CIFS on other large-scale skin datasets.